\documentclass{article} 
\usepackage{iclr2021_conference,times}
\usepackage[linesnumbered,ruled]{algorithm2e}

\iclrfinalcopy

\newtheorem{proposition}{Proposition}

\usepackage{graphicx} 
\usepackage{pdfpages}
\usepackage{caption}
\usepackage{subcaption}

\usepackage{amsmath,amsfonts,bm}

\def\eqref#1{equation~\ref{#1}}

\def\1{\bm{1}}

\DeclareMathAlphabet{\mathsfit}{\encodingdefault}{\sfdefault}{m}{sl}
\SetMathAlphabet{\mathsfit}{bold}{\encodingdefault}{\sfdefault}{bx}{n}

\usepackage{hyperref}
\usepackage{url}

\title{Distributionally Robust \\ Semi-Supervised Learning Over Graphs}

\author{Alireza Sadeghi, Meng Ma, Bingcong Li, Georgios B. Giannakis  \\ 
Department of ECE and Digital Technology Center \\ 
University of Minnesota \\
Minneapolis, MN 55455, USA \\
\texttt{\{sadeg012, maxxx971, lixx5599, georgios\}@umn.edu}
}

\begin{document}

\maketitle

\begin{abstract}
Semi-supervised learning (SSL) over graph-structured data emerges in many network science applications. To efficiently manage learning over graphs, variants of graph neural networks (GNNs) have been developed recently. By succinctly encoding local graph structures and features of nodes, state-of-the-art GNNs can scale linearly with the size of graph. Despite their success in practice, most of existing methods are unable to handle graphs with uncertain nodal attributes. Specifically whenever mismatches between training and testing data distribution exists, these models fail in practice. Challenges also arise due to distributional uncertainties associated with data acquired by noisy measurements. In this context, a distributionally robust learning framework is developed, where the objective is to train models that exhibit quantifiable robustness against perturbations. The data distribution is considered unknown, but lies within a Wasserstein ball centered around empirical data distribution. A robust model is obtained by minimizing the worst expected loss over this ball. However, solving the emerging functional optimization problem is challenging, if not impossible. Advocating a strong duality condition, we develop a principled method that renders the problem tractable and efficiently solvable. Experiments assess the performance of the proposed method.
\end{abstract}

\section{Introduction}
Relations among data in real world applications can often be captured by graphs, for instance the analysis and inference tasks for social, brain, communication, biological, transportation, and sensor networks \citep{shuman2013mag, kolaczyk2014statistical}. In practice however, the data is only available for a subset of nodes, due to for example the cost, and computational or privacy constraints. Most of these applications however, deal with inference of processes across all the network nodes. Such semi-supervised learning (SSL) tasks over networks can be addressed by exploiting the underlying graph topology \citep{sslbook}.

Graph neural networks (GNNs) are parametric models 
that combine graph-filters and topology information with point-wise nonlinearities, to form nested architectures to easily express the functions defined over graphs \citep{gnn2018survey}. By exploiting the underlying irregular structure of network data, the GNNs enjoy lower computational complexity, less parameters for training, and improved generalization capabilities relative to traditional deep neural networks (DNNs), making them appealing for learning over graphs \citep{gnn2018survey, gnn2020comprehensivesurvey, antonio2019tsp}.

Similar to other DNN models, GNNs are also susceptible to adversarial manipulated input data  or, distributional uncertainties, such as mismatches between training and testing data distributions. For instance small perturbations to input data would significantly deteriorate the regression performance, or result in classification error \citep{gnn2020adversarial, gnn2020adversarialsurvey}, just to name a few. Hence, it is critical to develop principled methods that can endow GNNs with robustness, especially in safety-critical applications, such as robotics \citep{multirobot2020learning}, and transportation \citep{gnn2020transport}.

\noindent \textbf{Contributions.} This paper endows SSL over graphs using GNNs with \textit{robustness} against distributional uncertainties and possibly adversarial perturbations. Assuming the data distribution lies inside a Wasserstein ball centered at empirical data distribution, we robustify the model by minimizing the worst expected loss over the considered ball, which is challenging to solve. Invoking recently developed strong duality results, we develop an equivalent unconstrained and tractable learning problem.

\subsection{Problem formulation}
Consider a SSL task over a graph $\mathcal{G} := \left\{\mathcal{V}, \mathbf{W} \right\}$ with $N$ nodes, where $\mathcal {V}:= \{1, \ldots, N\}$ denotes the vertex set, and $\mathbf{W}$ represents the $N\times N$ weighted adjacency matrix capturing node connectivity. The associated unnormalized graph Laplacian matrix of the undirected graph $\mathcal{G}$ is $\mathbf{L} := \mathbf{D} - \mathbf{A}$, where $\mathbf{D} := \textrm{diag}\{\mathbf{W} \mathbf{1}_N\}$, with $\mathbf{1}_N$ denoting the $N \times 1$ all-one column vector. Denote by matrix $\mathbf{X}_s \in \mathbb{R}^{N \times F}$ the nodal feature vectors sampled at instances $s = 1, 2, \cdots $, with $n$-th row $\mathbf{x}^\top_{n,s} := \left[\mathbf{X}_s\right]_{n:}$ representing a feature vector of length $F$ associated with node $n \in \mathcal{V}$, and $\top$ stands for transposition. 
In the given graph, the labels $\{y_{n,s}\}_{n \in \mathcal{O}_s}$ are given for \emph{only a small subset} of nodes,
where $\mathcal O_s$ represents the index set of \textit{observed} nodes sampled at~$s$, and $\mathcal{U}_s$ the index set of \textit{unobserved} nodes. 

Given $\{\mathbf{X}_s, \mathbf{y}_s\}$, where $\mathbf{y}_s$ is the vector of observed labels, the goal is to find the labels of unobserved nodes $\{y_{n,s}\}_{n \in \mathcal{U}_s}$. 
To this aim our objective is to learn a functional mapping $f(\mathbf{X}_s; \mathbf{W})$ that can infer the missing labels based on available information. 
Such a function can be learned by solving the following optimization problem (see e.g., \citet{kipf2016semi} for more details)
\begin{align}
\label{eq:loss}
\min_{f \in{\mathcal{F}}}  \mathbb{E}  \, \Big[   \overbrace{\underset{n \in \mathcal{O}_s}{\sum}  \|f(\mathbf{x}_{n}; \mathbf{W}) - y_{n}\|^2}^{\mathcal{L}_0} +  \lambda \overbrace{\underset{{n,n'}}{\sum} \mathbf{W}_{nn'} \| f(\mathbf{x}_n; \mathbf{W}) - f(\mathbf{x}_{n'}; \mathbf{W}) \|^2}^{\mathcal{L}_{\rm{reg}}} \Big],
\end{align}
where $\mathcal{L}_0$ represents the supervised loss w.r.t. the observed part of the graph, $\mathcal{L}_{\rm{reg}}$ represents the Laplacian regularization term, $\mathcal{F}$ denotes the feasible set of functions that we can learn, and $\lambda \ge 0$ is a hyper parameter. The regularization term relies on the premise that connected nodes in the graph are likely to share similar labels. The expectation here is taken with respect to (w.r.t) the feature and label data generating distribution.


In this work, we first encode the graph structure using a GNN model denoted by $f(\mathbf{X}; \boldsymbol{\theta}, \mathbf{W})$, where $\boldsymbol{\theta}$ represents the model parameters. Such a parametric representation enables bypassing explicit graph-based
regularization $\mathcal{L}_{\rm{reg}}$ represented in  \ref{eq:loss}. The GNN model of $f(\cdot)$ relies on the weighted adjacency $\mathbf{W}$ and therefore can easily propagate information from observed nodes $\mathcal{O}_s$ to unobserved ones $\mathcal{U}_s$. In a nutshell, objective is to learn a parametric model by solving the following problem
\begin{equation}
\underset{\boldsymbol{\theta} \in {\Theta}}{\min} \;\; \mathbb{E}_{\small{{\mathbf{\{X, y\}}\sim P_0}}} \; {\mathcal{L}_0\big(f(\mathbf{X}, \boldsymbol{\theta};\mathbf{W}), \mathbf{y}\big)} 
\end{equation} 
where $\Theta$ is a feasible set, and $P_0$ is the feature and label data generating distribution. Despite restricting the modeling capacity through parameterizing $f(\cdot)$ with GNNs, we may infuse additional prior information into the sought formulation through exploiting the weighted adjacency matrix $\mathbf{W}$, which does not necessarily encode node similarities. 

In practice, $P_0$ is typically unknown, instead some data samples $\{\mathbf{X}_s, \mathbf{y}_s\}_{s=1}^S$  are given. Upon replacing the nominal distribution with an empirical one, we arrive at the empirical loss minimization problem, that is $
{\min}_{\boldsymbol{\theta} \in {\Theta}} \;S^{-1} \sum_{s=1}^{S} \; {\mathcal{L}_0 \big(f(\mathbf{X}_s, \boldsymbol{\theta};\mathbf{W}), \mathbf{y}_s\big)}$. The model obtained by solving empirical risk minimization does not exhibit any robustness in practice, specifically if there is any mismatch between the training and testing data distributions. To endow robustness, we reformulate this learning problem in a fresh manner as described in ensuing section.

\section{Distributionally robust learning}
\label{gen_inst}
To endow robustness, we consider the following optimization problem
\begin{equation} \label{eq:robusterm}
\underset{\boldsymbol{\theta} \in {\Theta}}{\min} \; \underset{P \in \mathcal{P}}{\sup} \; \; \mathbb{E}_{\small{{\mathbf{(X, y)}\sim P}}} \; {\mathcal{L}_0\big(f(\mathbf{X}, \boldsymbol{\theta};\mathbf{W}), \mathbf{y}\big)} 
\end{equation} 
where $\mathcal P$ is a set of distributions centered around the \textit{empirical data distribution} $\widehat{P}_0$. This novel reformulation in \ref{eq:robusterm} yields a model that performs reasonably well among a continuum of distributions. Various ambiguity sets $\mathcal{P}$ can be considered in practice, and they lead to different robustness guarantees with different computational requirements. For instance momentum, KL divergence,  statistical test, and Wasserstein distance-based sets are popular in practice; see also \citep{blanchet2019quantifying, sinha2017certifying, blanchet2017data} and references therein. Among possible choices, KL divergence is not symmetric and under certain conditions can even become infinite, momentum based methods on the other hand are oftentimes not tractable in practice. Hence, here  we advocate the optimal transport theory and the Wasserstein distance to characterize the ambiguity set $\mathcal{P}$. As a result, we can offer a tractable solution for this problem, as delineated next.


To formalize our framework, let us first define the Wasserstein distance between two probability measures. To this aim, consider probability measures $P$ and $\widehat{P}$ supported on some set $\mathcal{X}$, and let $\Pi(P,\widehat{P})$ denote the set of joint measures (a.k.a coupling) defined over $\mathcal{X} \times \mathcal{X}$, with marginals $P$ and $\widehat{P}$, and let $c: \mathcal{X} \times \mathcal{X}  \rightarrow [0, \infty)$ measure the transportation cost for a unit of mass from $\mathbf{X \in \mathcal{X}}$ in $P$ to $\mathbf{X}' \in \mathcal{X}$ in $\widehat{P}$. The so-called optimal transport problem is concerned with the minimum cost associated with transporting all the mass from $P$ to $\widehat{P}$ through finding the optimal coupling, i.e., $
W_c(P,\widehat{P}) := \; \underset{\pi \in \Pi}{\inf} \, \mathbb{E}_\pi [ c(\mathbf {X},\mathbf{X}')]
$. If $c(\cdot,\cdot)$ satisfies the axioms of distance, then $W_c$ defines a distance on the space of probability measures. For instance, if $P$ and $\widehat{P}$ are defined over a Polish space equipped with metric $d$, then fixing $c(\mathbf X, \mathbf X') = d^p(\mathbf X, \mathbf X')$ for some $p\in [1, \infty)$ asserts that $W_c^{1/p}(P,\widehat{P})$ is the well-known Wasserstein distance of order $p$ between $P$ and $\widehat{P}$. 

Using the Wasserstein distance, let us define the uncertainty set $\mathcal{P} := \{P| W_c(P,\widehat{P}_0) \le \rho\}$ to include all probability distribution functions (pdfs) having at most $\rho$-distance from $\widehat{P}_0$. Incorporating this ambiguity set into \ref{eq:robusterm}, the following robust surrogate is considered in this work
\begin{equation}
	\label{eq:robform}
	\underset{\boldsymbol{\theta} \in {\Theta}}{\min}  \; \underset{P \in \mathcal{P}}{\sup} \quad \mathbb{E}_{\small{{\mathbf{(X, y)}\sim P}}} \; {\mathcal{L}_0\big(f(\mathbf{X}, \boldsymbol{\theta};\mathbf{W}), \mathbf{y}\big)}, \quad 
	{\rm where}\; \mathcal{P}:= \left\{P\, | W_c(P,\widehat{P}_0) \le \rho \right\}. 
\end{equation}
The inner supremum here goes after pdfs characterized by $\mathcal{P}$. Solving this optimization directly over the infinite-dimensional space of distribution functions raises practical challenges. Fortunately, under some mild conditions over losses as well as transport costs, the inner maximization satisfies a strong duality condition (see \citet{blanchet2017data} for a detailed discussions), which means the optimal objective of this inner maximization and its Lagrangian dual are equal. Enticingly, the dual reformulation involves optimization over only one-dimensional dual variable. These properties make it practically appealing to solve \ref{eq:robform} directly in the dual domain. The following proposition highlights the strong duality result, whose proofs can be found in \citep{blanchet2019quantifying}.  

\begin{proposition} \label{prop1}
	Under some mild conditions over the loss $\mathcal{L}_0(\cdot)$ and cost $c(\cdot)$, it holds that
	\begin{align} \label{eq:strongduality}
	\sup_{P\in\mathcal{P}} \, \mathbb{E}_{P} \; {\mathcal{L}_0\big(f(\mathbf{X}, \boldsymbol{\theta};\mathbf{W}), \mathbf{y}\big)}
	= \inf_{\gamma \ge 0}   \frac{1}{S} \sum_{s=1}^{S}    \sup_{\boldsymbol{\xi} \in {\mathcal X}}   \{ {\mathcal{L}_0\big(f(\bm{\xi}, \boldsymbol{\theta};\mathbf{W}), \mathbf{y}_s\big)} + \gamma \, (\rho - c(\mathbf X_s, \boldsymbol{\xi}) ) \}   
	\end{align}
	where $\mathcal{P}:= \left\{P\, | W_c(P,\widehat{P}_0) \le \rho \right\}$.
\end{proposition}

The right-hand side in \ref{eq:strongduality} simply is the univariate dual reformulation of the primal problem represented in the left-hand side. Furthermore, different from the primal formulation, the expectation in the dual domain is replaced with the summation over available training data, rather than any $P \in \mathcal{P}$ that needs to be obtained by solving for the optimal $\pi \in \Pi$ to form $\mathcal{P}$. Because of these two properties, solving the dual problem is practically more appealing. Thus, hinging on Proposition \ref{prop1}, the following distributionally robust surrogate is considered in this work 
\begin{align} \label{eq:robustdual}
\min_{\boldsymbol{\theta}\in \Theta} \; \inf_{\gamma \ge 0}  \frac{1}{S} \sum_{s=1}^{S}   \sup_{\boldsymbol{\xi} \in {\mathcal X}}  \left\{ 
{\mathcal{L}_0\big(f(\boldsymbol{\xi}, \boldsymbol{\theta};\mathbf{W}), \mathbf{y}_s \big)}
+  \gamma (\rho - c(\mathbf X_s, \boldsymbol{\xi}))\right\} \!\! 
\end{align}

This problem requires the supremum to be solved separately for each sample $\mathbf{X}_s$, which cannot be handled through existing methods. 
Our approach to address this relies on the structure of this problem to \textit{iteratively} update parameters ${\boldsymbol{\bar \theta}}:=[{\boldsymbol \theta}^\top, \gamma]^\top$ and $\boldsymbol{\xi}$. Specifically, we rely on Danskin's theorem to first maximize over $\boldsymbol{\xi}$, which results in a differentiable function of ${\boldsymbol{\bar \theta}}$, and then minimize the objective w.r.t. ${\boldsymbol{\bar \theta}}$ using gradient descent. However, to guarantee convergence to a stationary point and utilize Danskin's theorem, we need to make sure the inner maximization admits a unique solution (singleton). By choosing a strongly convex transportation cost such as $c(\mathbf{X}, \boldsymbol{\xi}) :=\| \mathbf{X} - \boldsymbol{\xi} \|_F^2$, and by selecting $\gamma \in \Gamma :=\{\gamma| \gamma >\gamma_0\}$ with a large enough $\gamma_0$, we arrive at a strongly concave objective function for the maximization over $\boldsymbol \xi$.
Since $\gamma$ is the dual variable associated with the constraint in \ref{eq:robform}, having $\gamma \in \Gamma$ is tantamount to tuning $\rho$, which in turn \emph{controls} the level of robustness. Replacing $\gamma \ge 0$ in \ref{eq:robustdual} with $\gamma \in \Gamma$, our \emph{robust model} can be obtained as the solution of 
\begin{align}
\min_{\boldsymbol{\theta}\in \Theta}~\inf_{\gamma \in \Gamma}~ \frac{1}{S} \sum_{s=1}^{S}  \sup_{\boldsymbol{\xi} \in\mathcal{X}} \psi(\mathbf{\boldsymbol{\bar{\theta}}}, \boldsymbol{\xi}; \mathbf{X}_s)   
\label{eq:objective}
\end{align}
where $ \psi(\mathbf{\boldsymbol{\bar{\theta}}}, \boldsymbol{\xi}; \mathbf{X}_s) = {\mathcal{L}_0(f(\boldsymbol{\xi}, \boldsymbol{\theta};\mathbf{W}), \mathbf{y}_s)} +\gamma (\rho - c(\mathbf X_s, \boldsymbol{\xi}))$.
Intuitively, input $\mathbf{X}_s$ in \ref{eq:objective} is pre-processed by maximizing $\psi(\cdot)$ accounting for a perturbation. We iteratively solve \ref{eq:objective}, where after sampling a mini-batch of data, we first pre-process them by maximizing the function $\psi(\cdot)$. Then, we use a simple gradient descent to update $\boldsymbol{\bar{\theta}}$. Notice that the $\bm{\theta}$ inside function $\psi(\cdot)$, represents the weights of our considered GNN, whose details are provided in the Appendix. A promising future research direction is to optimally/adaptively tune the  hyper parameter $\rho$. We refer interested readers to \cite{esfahani2018data, fournier2015rate} for further discussions. 

%
%

\begin{figure*}
	\begin{subfigure}[t]{0.24\textwidth}
		\centering
		\includegraphics[width=\textwidth]{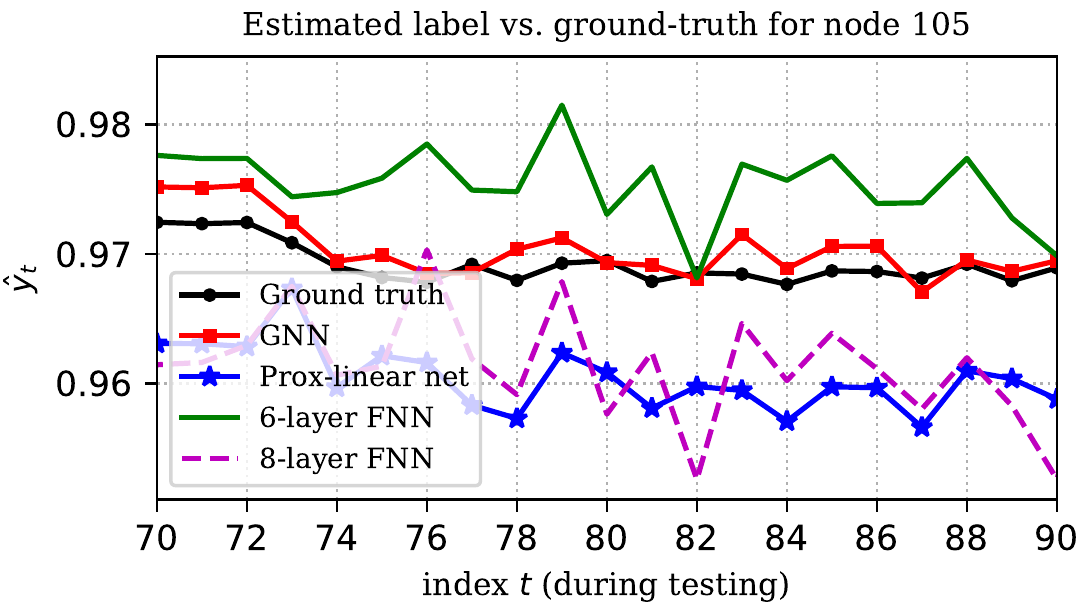}
		\caption{}
		\label{f  x}
	\end{subfigure}
	\hfill
	\begin{subfigure}[t]{0.24\textwidth}
		\centering
		\includegraphics[width=\textwidth]{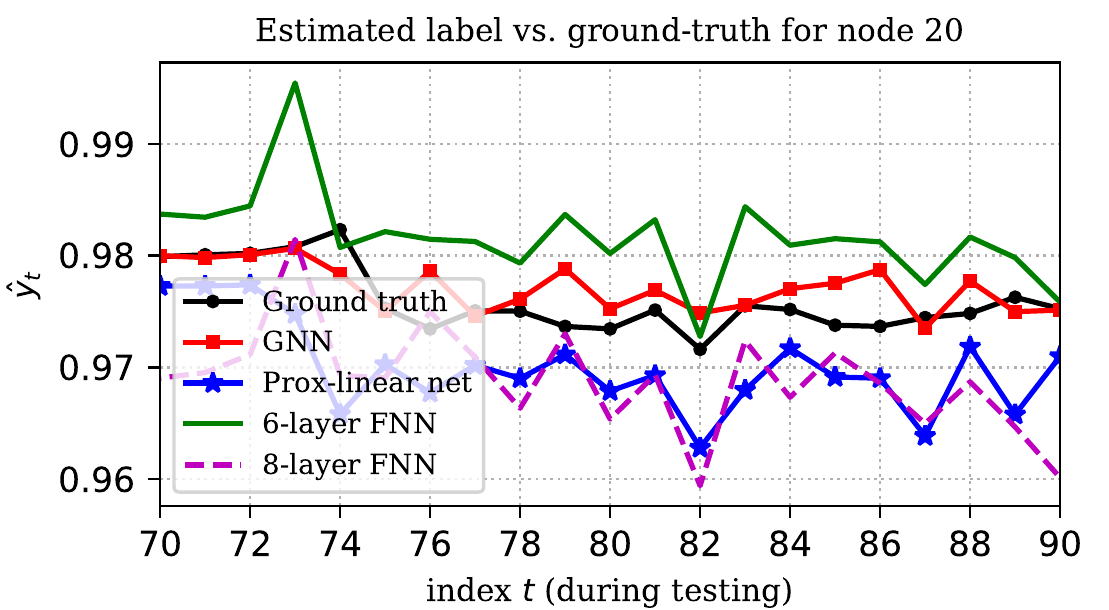}
		\caption{ }
		\label{fig:}
	\end{subfigure}
	\hfill
	\label{fig:}
	\begin{subfigure}[t]{0.24\textwidth}
		\centering
		\includegraphics[width=\textwidth]{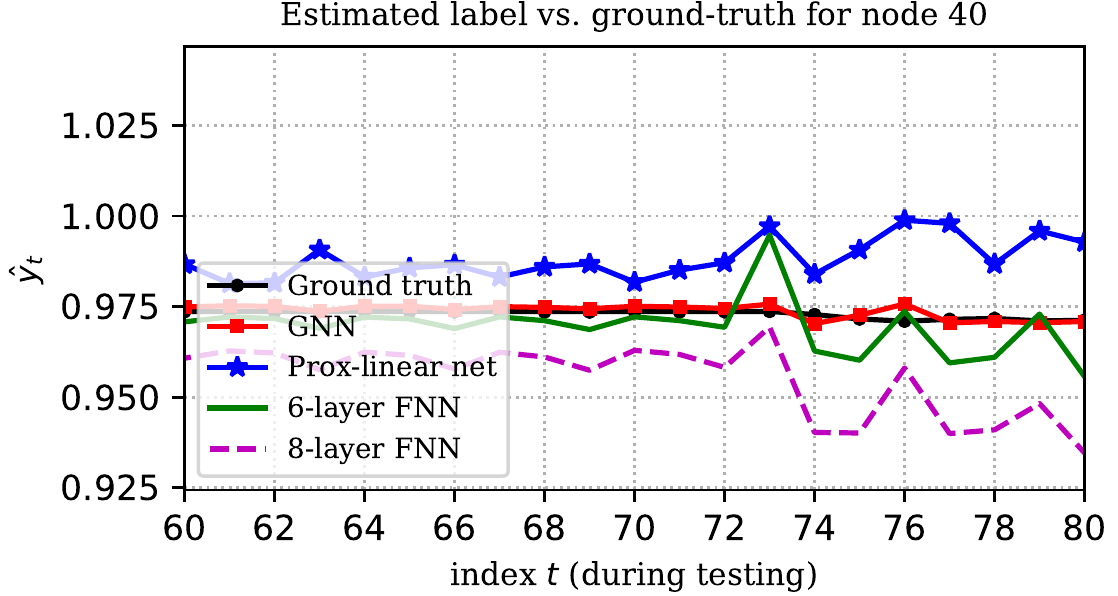}
		\caption{ }
		\label{fig:}
	\end{subfigure}
	\hfill
	\begin{subfigure}[t]{0.24\textwidth}
		\centering
		\includegraphics[width=\textwidth]{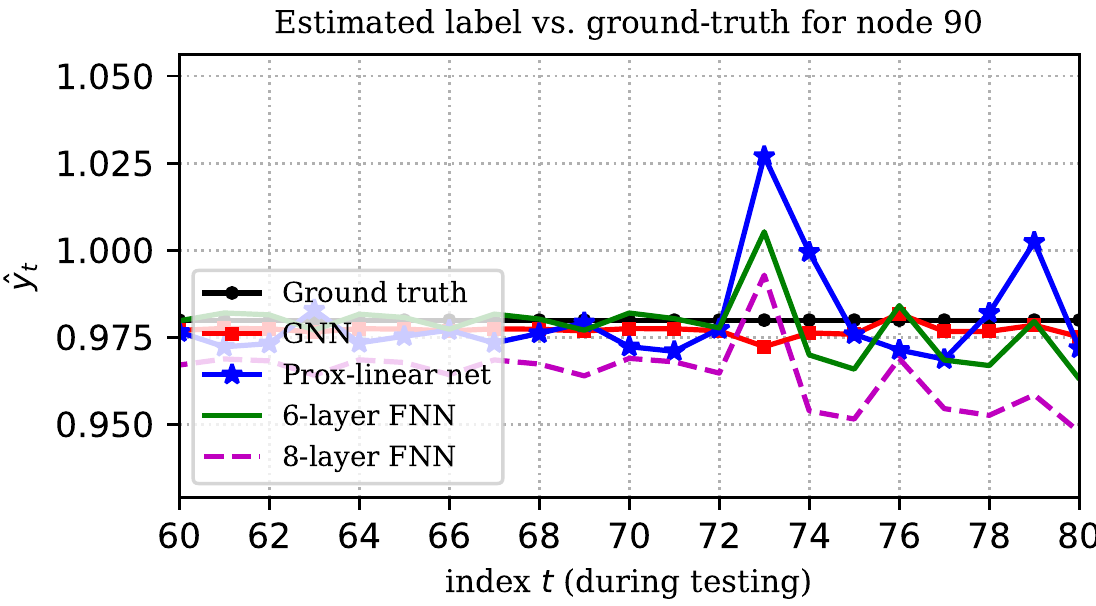}
		\caption{ }
		\label{fig:}
	\end{subfigure}
	\caption{Performance during testing for both normal (a) - (b), and perturbed input features (c) - (d).}
	\label{fig1}
\end{figure*}

\vspace{-.25 cm}
\section{Experiments}
\vspace{-.25 cm}
The performance of our novel distributionally robust GNN-based SSL is tested in a regression task using real 
load consumption data from the 2012 Global Energy Forecasting Competition (GEFC). Our objective here is to estimate only the amplitudes of voltages across all the nodes in a standard IEEE $118$-bus network.  Utilizing this data set, the training and testing data are prepared by solving the so-called AC power ﬂow equations using the MATPOWER toolbox \citep{matpower}. 

The measurements $\mathbf{X}$ used include all active and reactive power injections, corrupted by small additive white Gaussian noise. Using MATPOWER we generated $1,000$ pairs of measurements and ground-truth voltages. We used 80\% of this data for training and the remaining  for testing. Throughout the training, the Adam optimizer with a fixed learning rate $10^{-3}$ was employed to minimize the H\"uber loss. Furthermore, the batch size was set to $32$ during all $100$ epochs. 

To compare our method we employed $3$ different benchmarks, namely: i) the prox-linear network introduced in \citep{liang2019}; ii) a 6-layer vanilla feed-forward neural network (FNN); and, iii) an 8-layer FNN. Our considered GNN uses $K=2$ with $D=8$ hidden units with ReLU activation. 

The first set of tests are carried out using normal (not-corrupted) data, where the results are depicted in Fig. \ref{fig1}. Here we show the estimated (normalized) voltage amplitudes at different nodes, namely $105$, and $20$ during the given time course. The black curve represents the ground truth signal to be estimated. Clearly our GNN-based method outperforms alternative methods. 

The second set of experiments are carried out over corrupted input signals, and the results are reported in Fig. \ref{fig1}. Specifically the training samples were generated according to $P_0$, but during testing samples were perturbed to satisfy the constraint $P \in \mathcal{P}$, that would yield the worst expected loss. Fig. \ref{fig1} depicts  the estimated signals across nodes $40$ and $90$. Here we fixed $\rho = 10$ and related hyper-parameters are tuned using grid search. As the plots showcase, the our proposed GNN-based robust method outperforms competing alternatives with corrupted inputs.   

\vspace{-.25 cm}
\section{conclusions}
\vspace{-.25 cm}
This contribution dealt with semi-supervised learning over graphs using GNNs. To account for uncertainties associated with data distributions, or adversarially manipulated input data, a principled robust learning framework was developed. Using the parametric models, we were able to reconstruct the unobserved nodal values. 

\newpage

\bibliography{iclr2021_conference}

\begin{thebibliography}{18}
\providecommand{\natexlab}[1]{#1}
\providecommand{\url}[1]{\texttt{#1}}
\expandafter\ifx\csname urlstyle\endcsname\relax
  \providecommand{\doi}[1]{doi: #1}\else
  \providecommand{\doi}{doi: \begingroup \urlstyle{rm}\Url}\fi

\bibitem[Blanchet \& Murthy(2019)Blanchet and Murthy]{blanchet2019quantifying}
J.~Blanchet and K.~Murthy.
\newblock Quantifying distributional model risk via optimal transport.
\newblock \emph{Math. Oper. Res.}, 44:\penalty0 565--600, 2019.

\bibitem[Blanchet et~al.(2006)Blanchet, Kang, Zhang, and
  Murthy]{blanchet2017data}
J.~Blanchet, Y.~Kang, F.~Zhang, and K.~Murthy.
\newblock Data-driven optimal transport cost selection for distributionally
  robust optimization.
\newblock \emph{Stat.}, 1050:\penalty0 1527--1554, 2006.

\bibitem[Chapelle et~al.(2009)Chapelle, Scholkopf, and Zien]{sslbook}
O.~Chapelle, B.~Scholkopf, and A.~Zien.
\newblock Semi-supervised learning.
\newblock \emph{IEEE Trans. Neural Netw.}, 3:\penalty0 542, 2009.

\bibitem[Esfahani \& Kuhn(2018)Esfahani and Kuhn]{esfahani2018data}
P.~M. Esfahani and D.~Kuhn.
\newblock Data-driven distributionally robust optimization using the
  {Wasserstein metric: Performance} guarantees and tractable reformulations.
\newblock \emph{Math. Prog.}, 171\penalty0 (1):\penalty0 115--166, 2018.

\bibitem[Fournier \& Guillin(2015)Fournier and Guillin]{fournier2015rate}
N.~Fournier and A.~Guillin.
\newblock On the rate of convergence in {Wasserstein} distance of the empirical
  measure.
\newblock \emph{Probability Theory and Related Fields}, 162\penalty0
  (3):\penalty0 707--738, 2015.

\bibitem[{Gama} et~al.(2019){Gama}, {Marques}, {Leus}, and
  {Ribeiro}]{antonio2019tsp}
F.~{Gama}, A.~G. {Marques}, G.~{Leus}, and A.~{Ribeiro}.
\newblock Convolutional neural network architectures for signals supported on
  graphs.
\newblock \emph{IEEE Trans. Signal Process.}, 67:\penalty0 1034--1049, 2019.

\bibitem[Jin et~al.(2020)Jin, Li, Xu, Wang, and Tang]{gnn2020adversarialsurvey}
W.~Jin, Y.~Li, H.~Xu, Y.~Wang, and J.~Tang.
\newblock Adversarial attacks and defenses on graphs: A review and empirical
  study.
\newblock \emph{arXiv:2003.00653}, 2020.

\bibitem[Kipf \& Welling(2016)Kipf and Welling]{kipf2016semi}
T.~N. Kipf and M.~Welling.
\newblock Semi-supervised classification with graph convolutional networks.
\newblock \emph{Intl. Conf. Lear. Rep.}, 2016.

\bibitem[Kolaczyk \& Cs{\'a}rdi(2014)Kolaczyk and
  Cs{\'a}rdi]{kolaczyk2014statistical}
E.~D. Kolaczyk and G.~Cs{\'a}rdi.
\newblock \emph{Statistical analysis of network data with R}, volume~65.
\newblock Springer, 2014.

\bibitem[{Shuman} et~al.(2013){Shuman}, {Narang}, {Frossard}, {Ortega}, and
  {Vandergheynst}]{shuman2013mag}
D.~I. {Shuman}, S.~K. {Narang}, P.~{Frossard}, A.~{Ortega}, and
  P.~{Vandergheynst}.
\newblock The emerging field of signal processing on graphs: Extending
  high-dimensional data analysis to networks and other irregular domains.
\newblock \emph{IEEE Signal Proces. Mag.}, 30:\penalty0 83--98, 2013.

\bibitem[Sinha et~al.(2017)Sinha, Namkoong, and Duchi]{sinha2017certifying}
A.~Sinha, H.~Namkoong, and R.~Volpiand~J. Duchi.
\newblock Certifying some distributional robustness with principled adversarial
  training.
\newblock \emph{Intl. Conf. Learn. Rep.}, 2017.

\bibitem[Tolstaya et~al.(2020)Tolstaya, Gama, Paulos, Pappas, Kumar, and
  Ribeiro]{multirobot2020learning}
E.~Tolstaya, F.~Gama, J.~Paulos, G.~Pappas, V.~Kumar, and A.~Ribeiro.
\newblock Learning decentralized controllers for robot swarms with graph neural
  networks.
\newblock In \emph{Conf. Robot Learn.}, pp.\  671--682, 2020.

\bibitem[Wu et~al.(2020)Wu, Pan, Chen, Long, Zhang, and
  Philip]{gnn2020comprehensivesurvey}
Z.~Wu, S.~Pan, F.~Chen, G.~Long, C.~Zhang, and S.~Yu Philip.
\newblock A comprehensive survey on graph neural networks.
\newblock \emph{IEEE Trans. Neural Netw. Learn. Syst.}, 2020.

\bibitem[{Zhang} et~al.(2019){Zhang}, {Wang}, and {Giannakis}]{liang2019}
L.~{Zhang}, G.~{Wang}, and G.~B. {Giannakis}.
\newblock Real-time power system state estimation and forecasting via deep
  unrolled neural networks.
\newblock \emph{IEEE Trans. on Signal Proces.}, 67\penalty0 (15):\penalty0
  4069--4077, 2019.

\bibitem[{Zhou} et~al.(2020){Zhou}, {Yang}, {Zhong}, {Chen}, and
  {Zhang}]{gnn2020transport}
F.~{Zhou}, Q.~{Yang}, T.~{Zhong}, D.~{Chen}, and N.~{Zhang}.
\newblock Variational graph neural networks for road traffic prediction in
  intelligent transportation systems.
\newblock \emph{IEEE Trans. Ind. Inf.}, pp.\  2802--2812, 2020.

\bibitem[Zhou et~al.(2018)Zhou, Cui, Zhang, Yang, Liu, Wang, Li, and
  Sun]{gnn2018survey}
J.~Zhou, G.~Cui, Z.~Zhang, C.~Yang, Z.~Liu, L.~Wang, C.~Li, and M.~Sun.
\newblock Graph neural networks: A review of methods and applications.
\newblock \emph{arXiv:1812.08434}, 2018.

\bibitem[Zimmerman et~al.(2010)Zimmerman, Murillo-S{\'a}nchez, and
  Thomas]{matpower}
R.~D. Zimmerman, C.~E. Murillo-S{\'a}nchez, and R.~J. Thomas.
\newblock {MATPOWER: Steady-state} operations, planning, and analysis tools for
  power systems research and education.
\newblock \emph{IEEE Trans. Power syst.}, 26:\penalty0 12--19, 2010.

\bibitem[Z{\"u}gner et~al.(2020)Z{\"u}gner, Borchert, Akbarnejad, and
  Guennemann]{gnn2020adversarial}
D.~Z{\"u}gner, O.~Borchert, A.~Akbarnejad, and S.~Guennemann.
\newblock Adversarial attacks on graph neural networks: Perturbations and their
  patterns.
\newblock \emph{ACM Trans. Knwl. Discov. from Data}, 14:\penalty0 1--31, 2020.

\end{thebibliography}
\bibliographystyle{iclr2021_conference}

\appendix
\section{Appendix--Graph neural networks}
GNNs are parametric models to represent functional  relationship for graph structured data. Specifically, the input to a GNN is a data matrix $\mathbf X$. Upon multiplying the input $\mathbf X$ by $\mathbf W$, features will diffuse over the graph, giving a new graph signal $ \check{\mathbf{Y}}=\mathbf{W X}$. To model feature propagation, one can also replace $\mathbf{W}$ with the (normalized) graph Laplacian or random walk Laplacian, since they will also preserve dependencies among nodal attributes.

During the diffusion process, the feature vector of each node is updated by a linear combination of its neighbors. Take the $n$-th node as an example, the shifted $f$-th feature $[\check{\mathbf Y}]_{nf}$ is obtained by $ [\check{\mathbf{Y}}]_{nf} = \sum_{i=1}^{N}[\mathbf{W}]_{ni}[\mathbf{X}]_{if}=\sum_{i \in \mathcal{N}_{n}} w_{ni} x_{i}^{f} $, 
where ${\mathcal{N}}_{n}$ denotes the set of neighboring nodes for node $n$. 
The so-called convolution operation in GNNs utilizes topology to combine features, namely
\begin{equation}
\label{eq:gc}
[\mathbf{Y}]_{nd}:=[\mathcal{H} \star \mathbf{X}; \mathbf{W}]_{nd}:=\sum_{k=0}^{K-1}[\mathbf{W}^{k} \mathbf{X}]_{n:} [\mathbf{H}_k]_{:d} 
\end{equation}
where $\mathcal{H}:=[\mathbf{H}_0~ \cdots~\mathbf{H}_{K-1}]$ with ${\mathbf H}_{k} \in \mathbb{R}^{F\times D}$ as filter coefficients;~$\mathbf Y \in \mathbb R^{N\times D}$ the intermediate (hidden) matrix with $D$ features per node;~and $\mathbf{W}^{k} \mathbf{X}$ as the linearly combined features of nodes within the $k$-hop neighborhood.

To construct a GNN with $L$ hidden layers, first let us denote by $\mathbf{X}_{l-1}$ the output of the $(l-1)$-th layer, which is also the $l$-th layer input for $l=1, \ldots, L$, and $\mathbf{X}_0 = \mathbf{X}$ to represent the input matrix. The hidden $\mathbf{Y}_{l} \in \mathbb{R}^{N\times D_{l}}$ with $D_l$ features is obtained by applying the graph convolution operation \ref{eq:gc} at layer $l$, i.e., $[\mathbf{Y}_l]_{nd} =\sum_{k=0}^{K_l-1}[\mathbf{W}^{k} \mathbf{X}_{l-1}]_{n:} [\mathbf{H}_{lk}]_{:g}$, 
where $\mathbf{H}_{lk} \in \mathbb{R}^{F_{l-1}\times F_{l}}$ is the convolution coefficients for $k=0, \ldots, K_l-1$. The output at layer $l$ is constructed by applying a graph convolution followed by a point-wise nonlinear operation $\sigma_{l}(\cdot)$. The input-output relationship at layer $l$ can be represented succinctly by $
\mathbf{X}_{l} =\sigma_{l}(\mathbf Y_{l})=\sigma_{l}\!\left(\sum_{k=0}^{K_l-1} \mathbf{W}^{k} \mathbf{X}_{l-1} \mathbf{H}_{l k}\right).
$ 
Using this mapping, GNNs use a nested architecture to represent nonlinear functional operator $\mathbf{X}_L={f}(\mathbf{X}_{0} ; \bm{\theta}, \mathbf{W})$ that maps the GNN input $\mathbf{X}_{0}$ to label estimates by taking into account the graph structure through $\mathbf W$. Specifically, in a compact representation we have that
\begin{align}  \label{eq:GNNbasepsse} 
 f(\mathbf{X}_{0} ; \bm{\theta}, \mathbf{W}) :=  
 \sigma_{L}\!\left(\sum_{k=0}^{K_{L }-1} \mathbf{W}^{k} \!\left(\ldots \!\left(\sigma_{1}\!\left(\sum_{k=0}^{K_1-1} \mathbf{W}^{k} \mathbf{X}_{0} \mathbf{H}_{1 k}\right) \ldots\right)\right)\mathbf{H}_{L k}\right)  
\end{align}
where the parameter set $\bm \theta$ contains all the \textit{trainable} filter weights $\{\mathbf{H}_{lk}, \forall l, k\}$.

\end{document}